\documentclass[10pt,twocolumn,letterpaper]{article}
\pdfoutput=1
\usepackage{cvpr}
\usepackage{times}
\usepackage{epsfig}
\usepackage{graphicx}
\usepackage{amsmath}
\usepackage{amssymb}
\usepackage{multirow}
\usepackage{subfigure}
\usepackage{bm}
\usepackage{xcolor}
\usepackage{threeparttable}
\usepackage{authblk}

\usepackage[pagebackref=true,breaklinks=true,letterpaper=true,colorlinks,bookmarks=false]{hyperref}

\renewcommand{\mathbf}{\boldsymbol}
\newcommand{\vv}{\boldsymbol v}
\newcommand{\h}{\boldsymbol h}
\newcommand{\w}{\boldsymbol w}
\newcommand{\x}{\boldsymbol x}
\newcommand{\y}{\boldsymbol y}
\newcommand{\p}{\boldsymbol p}

\newcommand{\E}{\boldsymbol E}
\newcommand{\V}{\boldsymbol V}
\newcommand{\U}{\boldsymbol U}
\newcommand{\W}{\boldsymbol W}
\newcommand{\aalpha}{\boldsymbol \alpha}
\newcommand{\bbeta}{\boldsymbol \beta}

\cvprfinalcopy 

\def\cvprPaperID{518} 

\ifcvprfinal\fi
\begin{document}

\title{Image Captioning with Semantic Attention}

\author[1]{Quanzeng You}
\author[2]{Hailin Jin}
\author[2]{Zhaowen Wang}
\author[2]{Chen Fang}
\author[1]{Jiebo Luo}
\affil[1]{Department of Computer Science, University of Rochester, Rochester NY 14627, USA}
\affil[2]{Adobe Research, 345 Park Ave, San Jose CA 95110, USA  \authorcr
\tt\small{\{qyou,jluo\}@cs.rochester.edu}, \{hljin,zhawang,cfang\}@adobe.com}

\maketitle

\begin{abstract}
Automatically generating a natural language description of an image has
attracted interests recently both because of its importance in practical
applications and because it connects two major artificial intelligence
fields: computer vision and natural language processing. Existing
approaches are either top-down, which start from a gist of an image and
convert it into words, or bottom-up, which come up with words describing
various aspects of an image and then combine them. In this paper, we
propose a new algorithm that combines both approaches through a model of
semantic attention. Our algorithm learns to selectively attend to semantic
concept proposals and fuse them into hidden states and outputs of
recurrent neural networks. The selection and fusion form a feedback
connecting the top-down and bottom-up computation.  We evaluate our
algorithm on two public benchmarks: Microsoft COCO and Flickr30K. Experimental results show that our algorithm significantly
outperforms the state-of-the-art approaches consistently across
different evaluation metrics.
\end{abstract}

\vspace{-2mm}
\section{Introduction}
\label{sec:intro}

Automatically generating a natural language description of an image, a
problem known as image captioning, has recently received a lot of
attention in Computer Vision. The problem is interesting not only
because it has important practical applications, such as helping
visually impaired people see, but also because it is regarded as a grand
challenge for image understanding which is a core problem in Computer
Vision. Generating a meaningful natural language description of an image
requires a level of image understanding that goes well beyond image
classification and object detection. The problem is also interesting in
that it connects Computer Vision with Natural Language Processing which
are two major fields in Artificial Intelligence.
\begin{figure}[t!]
\begin{center}
\includegraphics[width=0.48\textwidth]{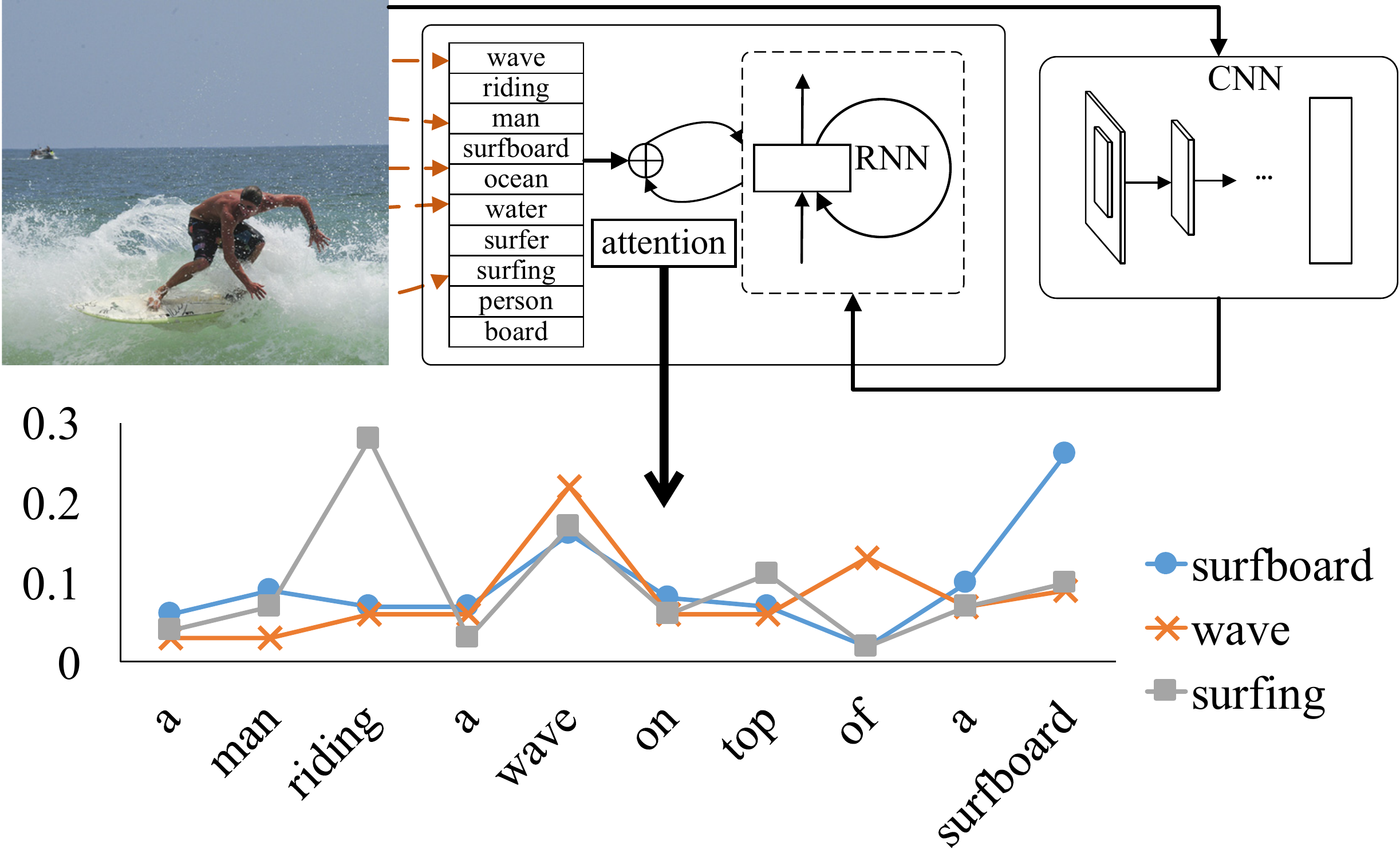}
\vspace{-1mm}
\caption{{\bf Top}: an overview of the proposed framework. Given an image, we use a
  convolutional neural network to extract a top-down visual feature and
  at the same time detect visual concepts (regions, objects,
  attributes, etc.). We employ a semantic attention model to combine the
  visual feature with visual concepts in a recurrent neural network that
  generates the image caption. {\bf Bottom}: We show the changes of the
  attention weights for several candidate concepts with respect to the
  recurrent neural network iterations.}
\label{fig:brief:intro}
\vspace{-6mm}
\end{center}
\end{figure}

There are two general paradigms in existing image captioning approaches:
top-down and bottom-up. The top-down paradigm \cite{chen2014learning,vinyals2014show,
  mao2014deep, karpathy2014deep, donahue2014long, xu2015show,
  mao2015learning} starts from a ``gist'' of an image and converts it
into words, while the bottom-up one \cite{farhadi2010every,
  kulkarni2011baby, li2011composing, elliott2013image, kuznetsova2012collective,
  fang2014captions, lebret2014simple} first comes up with words
describing various aspects of an image and then combines them. Language
models are employed in both paradigms to form coherent sentences. The
state-of-the-art is the top-down paradigm where there is an end-to-end
formulation from an image to a sentence based on recurrent neural
networks and all the parameters of the recurrent network can be
learned from training data. One of the limitations of the top-down
paradigm is that it is hard to attend to fine details which may be
important in terms of describing the image. Bottom-up approaches do not
suffer from this problem as they are free to operate on any image
resolution. However, they suffer from other problems such as there lacks
an end-to-end formulation for the process going from individual aspects
to sentences. There leaves an interesting question: Is it possible to
combine the advantages of these two paradigms? This naturally leads to
{\em feedback} which is the key to combine top-down and bottom-up
information.

Visual attention \cite{koch1987shifts, spratling2004feedback} is an important
mechanism in the visual system of primates and humans. It is a feedback
process that selectively maps a representation from the early stages in
the visual cortex into a more central non-topographic representation
that contains the properties of only particular regions or objects in
the scene. This selective mapping allows the brain to focus
computational resources on an object at a time, guided by low-level
image properties. The visual attention mechanism also plays an important
role in natural language descriptions of images biased towards
semantics. In particular, people do not describe everything in an
image. Instead, they tend to talk more about semantically more important
regions and objects in an image.

In this paper, we propose a new image captioning approach that combines
the top-down and bottom-up approaches through a semantic attention
model. Please refer to Figure~\ref{fig:brief:intro} for an overview of our
algorithm. Our definition for semantic attention in image captioning is
the ability to provide a detailed, coherent description of semantically
important objects that are needed exactly when they are needed. In
particular, our semantic attention model has the following properties:
1) able to attend to a semantically important concept or region of
interest in an image, 2) able to weight the relative strength of
attention paid on multiple concepts, and 3) able to switch attention
among concepts dynamically according to task status.  Specifically, we
detect semantic concepts or attributes as candidates for attention using
a bottom-up approach, and employ a top-down visual feature to guide
where and when attention should be activated.  Our model is built on top
of a Recurrent Neural Network (RNN), whose initial state
captures global information from the top-down feature.  As the RNN state
transits, it gets feedback and interaction from the bottom-up attributes
via an attention mechanism enforced on both network state and output
nodes. This feedback allows the algorithm to not only predict more
accurately new words, but also lead to more robust inference
of the semantic gap between existing predictions and image content.

\subsection{Main contributions}
\label{sec:contrib}

The main contribution of this paper is a new image captioning algorithm
that is based on a novel semantic attention model. Our attention model
naturally combines the visual information in both top-down and bottom-up
approaches in the framework of recurrent neural networks. Our algorithm
yields significantly better performance compared to the state-of-the-art
approaches. For instance, on Microsoft COCO and Flickr 30K, our
algorithm outperforms competing methods consistently across different evaluation metrics
(Bleu-1,2,3,4, Meteor, and Cider). We also conduct an extensive study to
compare different attribute detectors and attention schemes.

It is worth pointing out that \cite{xu2015show} also considered using
attention for image captioning. There are several important differences
between our work and \cite{xu2015show}. First, in \cite{xu2015show}
attention is modeled spatially at a fixed resolution. At every recurrent
iteration, the algorithm computes a set of attention weights
corresponding to pre-defined spatial locations. Instead, we can use
concepts from anywhere at any resolution in the image. Indeed, we can
even use concepts that do not have direct visual presence in the image.
Second, in our work there is a feedback process that combines top-down
information (the global visual feature) with bottom-up concepts which
does not exist in \cite{xu2015show}. Third, in \cite{xu2015show} uses
pretrained feature at a particular spatial location. Instead, we use
word features that correspond to detected visual concepts. This way, we
can leverage external image data for training visual concepts and
external text data for learning semantics between words.

\section{Related work}
\label{sec:related}

There is a growing body of literature on image captioning which can be
generally divided into two categories: top-down and bottom-up. Bottom-up
approaches are the ``classical'' ones, which start with visual concepts,
objects, attributes, words and phrases, and combine them into sentences
using language models. \cite{farhadi2010every} and
\cite{kulkarni2011baby} detect concepts and use templates to obtain
sentences, while \cite{li2011composing} pieces together detected
concepts. \cite{elliott2013image} and \cite{kuznetsova2012collective}
use more powerful language models. \cite{fang2014captions} and
\cite{lebret2014simple} are the latest attempts along this direction and
they achieve close to the state-of-the-art performance on
various image captioning benchmarks.

Top-down approaches are the ``modern'' ones, which formulate image
captioning as a machine translation problem \cite{sutskever2014sequence,
  bahdanau2014neural, cho2014learning}. Instead of translating between
different languages, these approaches translate from a visual
representation to a language counterpart. The visual representation
comes from a convolutional neural network which is often pretrained for
image classification on large-scale datasets \cite{krizhevsky2012imagenet}. Translation is
accomplished through recurrent neural networks based language
models. The main advantage of this approach is that
the entire system can be trained from end to end, i.e., all the
parameters can be learned from data.  Representative works
include \cite{vinyals2014show, mao2014deep, karpathy2014deep,
  donahue2014long, xu2015show, mao2015learning}.  The differences of the
various approaches often lie in what kind of recurrent neural networks
are used. Top-down approaches represent the state-of-the-art in this
problem.

Visual attention is known in Psychology and Neuroscience for long but is
only recently studied in Computer Vision and related areas. In terms of
models, \cite{larochelle2010learning, tang2014learning} approach it with Boltzmann
machines while \cite{mnih2014recurrent} does with recurrent neural
networks. In terms of applications, \cite{denil2012learning} studies it for image tracking,
\cite{ba2014multiple} studies it for image recognition of multiple
objects, and \cite{gregor2015draw} uses for image generation.  Finally, as
we discuss in Section~\ref{sec:intro}, we are not the first to consider it
for image captioning. In \cite{xu2015show}, Xu et al., propose a
spatial attention model for image captioning.
\section{Semantic attention for image captioning}
\label{sec:att}

\subsection{Overall framework}

We extract both top-down and bottom-up features from an input image.
First, we use the intermediate filer responses from a classification Convolutional Neural Network (CNN)
to build a global visual description denoted by $\vv$.
In addition, we run a set of attribute detectors to get a list of visual attributes or concepts $\{A_i\}$
that are most likely to appear in the image.
Each attribute $A_i$ corresponds to an entry in our vocabulary set or dictionary $\mathcal{Y}$.
The design of attribute detectors will be discussed in Section~\ref{sec:attribute}.

\begin{figure}[t]
\begin{center}
\includegraphics[width=0.95\linewidth]{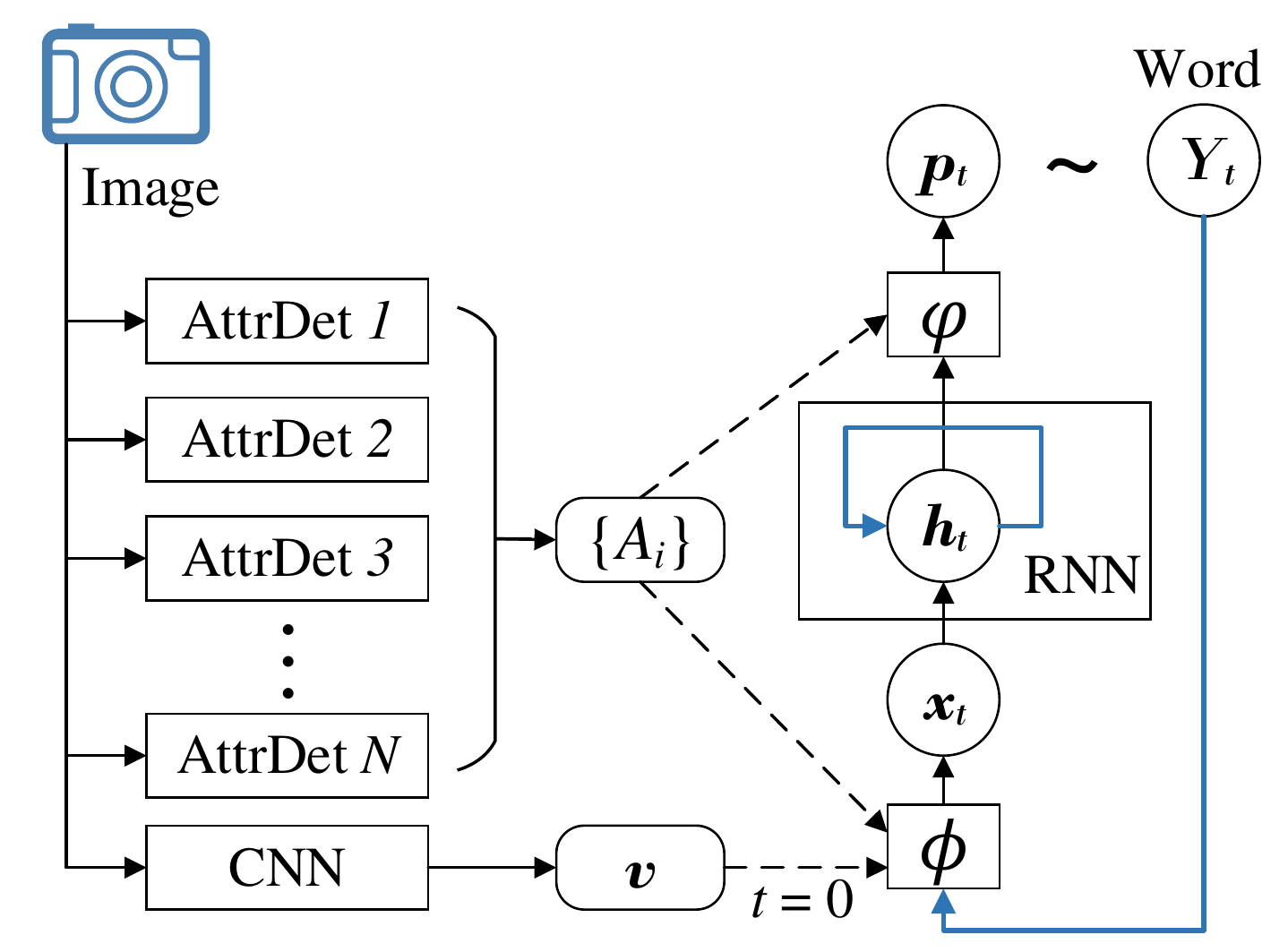}
\end{center}
\vspace{-1mm}
\caption{The framework of the proposed image captioning system. Visual features of CNN responses $\vv$ and attribute detections $\{A_i\}$
are injected into RNN (dashed arrows) and get fused together through a feedback loop (blue arrows).
Attention on attributes is enforced by both input model $\phi$ and output model $\varphi$.}
\label{fig:framework}
\vspace{-4mm}
\end{figure}

All the visual features are fed into a Recurrent Neural Network (RNN) for caption generation.
As the hidden state $\h_t \in \mathbb{R}^n$ in RNN evolves over time $t$, the $t$-th word $Y_t$ in the caption
is drawn from the dictionary $\mathcal{Y}$ according to a probability vector $\p_t \in \mathbb{R}^{|\mathcal{Y}|}$ controlled by the state $\h_t$.
The generated word $Y_t$ will be fed back into RNN in the next time step as part of the network input $\x_{t+1} \in \mathbb{R}^m$,
which drives the state transition from $\h_t$ to $\h_{t+1}$.
The visual information from $\vv$ and $\{A_i\}$ serves as as an external guide for RNN in generating $\x_t$ and $\p_t$,
which is specified by input and output models $\phi$ and $\varphi$.
The whole model architecture is illustrated in~\figurename~\ref{fig:framework}.

Different from previous image captioning methods, our model has a unique way to utilize and combine
different sources of visual information.
The CNN image feature $\vv$ is only used in the initial input node $\x_0$, which is expected to give RNN
a quick overview of the image content.
Once the RNN state is initialized to encompass the overall visual context, it is able to select specific items from $\{A_i\}$
for task-related processing in the subsequent time steps.
Specifically, the main working flow of our system is governed by the following equations:
\begin{align}
    \x_0 = &\; \phi_0(\vv) = \W^{x,v} \vv  \label{eqn:x0} \\
    \h_t = &\; \mbox{RNN}(\h_{t-1}, \x_t) \label{eqn:rnn} \\
    Y_t \sim &\; \p_t = \varphi(\h_t, \{A_i\}) \label{eqn:attout_abs} \\
    \x_t = &\; \phi(Y_{t-1}, \{A_i\}), \;\; t>0 , \label{eqn:attin_abs}
\end{align}
where a linear embedding model is used in Eq. \eqref{eqn:x0} with weight $\W^{x,v}$.
For conciseness, we omit all the bias terms of linear transformations in the paper.
The input and output attention models in Eq. \eqref{eqn:attout_abs} and \eqref{eqn:attin_abs} are designed to
adaptively attend to certain cognitive cues in $\{A_i\}$ based on the current model status,
so that the extracted visual information will be most relevant to the parsing of existing words
and the prediction of future word.
Eq.~\eqref{eqn:rnn} to \eqref{eqn:attin_abs} are recursively applied, through which
the attended attributes are fed back to state $\h_t$ and integrated with the global information from $\vv$.
The design of Eq. \eqref{eqn:attout_abs} and \eqref{eqn:attin_abs} is discussed below.

\subsection{Input attention model}
\label{sec:attin}

In the input attention model $\phi$ for $t{>}0$, a score $\alpha^i_t$ is assigned to each detected attribute $A_i$
based on its relevance with the previous predicted word $Y_{t-1}$.
Since both $Y_{t-1}$ and $A_i$ correspond to an entry in dictionary $\mathcal{Y}$,
they can be encoded with one-hot representations in $\mathbb{R}^{|\mathcal{Y}|}$ space,
which we denote as $\y_{t-1}$ and $\y^i$ respectively.
As a common approach to model relevance in vector space, a bilinear function is used to evaluate $\alpha^i_t$:
\begin{equation}
\label{eqn:alpha1}
    \alpha^i_t \propto \exp\big(\y_{t-1}^T \tilde{\U} \y^i\big) ,
\end{equation}
where the exponent is taken to normalize over all the $\{A_i\}$ in a softmax fashion.
The matrix $\tilde{\U} \in \mathbb{R}^{|\mathcal{Y}|\times|\mathcal{Y}|}$ contains a huge number of parameters for
any $\mathcal{Y}$ with a reasonable vocabulary size.
To reduce parameter size, we can first project the one-hot representations into a low dimensional word vector space
with Word2Vec \cite{mikolov2013distributed} or Glove \cite{pennington2014glove}.
Let the word embedding matrix be $\E \in \mathbb{R}^{d\times|\mathcal{Y}|}$ with $d\ll|\mathcal{Y}|$;
Eq.~\eqref{eqn:alpha1} becomes
\begin{equation}
\label{eqn:alpha2}
    \alpha^i_t \propto \exp\left(\y_{t-1}^T \E^T\U\E \y^i\right),
\end{equation}
where $\U$ is a $d \times d$ matrix.

Once calculated, the attention scores are used to modulate the strength of attention on different attributes.
The weighted sum of all attributes is mapped from word embedding space to the input space of $\x_t$
together with the previous word:
\begin{equation}
\label{eqn:attin}
    \x_t = \W^{x,Y} \big(\E\y_{t-1} + \mbox{diag}(\w^{x,A})\sum_i\alpha^i_t \E\y^i \big) ,
\end{equation}
where $\W^{x,Y} \in \mathbb{R}^{m \times d}$ is the projection matrix,
$\mbox{diag}(\w)$ denotes a diagonal matrix constructed with vector $\w$,
and $\w^{x,A}\in \mathbb{R}^d$ models the relative importance of visual attributes
in each dimension of the word space.

\subsection{Output attention model}
\label{sec:attout}

The output attention model $\varphi$ is designed similarly as the input attention model.
However, a different set of attention scores are calculated since visual concepts may be attended
in different orders during the analysis and synthesis processes of a single sentence.
With all the information useful for predicting $Y_t$ captured by the current state $\h_t$,
the score $\beta_t^i$ for each attribute $A_i$ is measured with respect to $\h_t$:
\begin{equation}
\label{eqn:beta}
    \beta^i_t \propto \exp\left(\h_t^T \V \sigma(\E\y^i)\right),
\end{equation}
where $\V \in \mathbb{R}^{n \times d}$ is the bilinear parameter matrix.
$\sigma$ denotes the activation function connecting input node to hidden state in RNN,
which is used here to ensure the same nonlinear transform is applied to the two feature vectors before they are compared.

Again, $\{\beta^i_t\}$ are used to modulate the attention on all the attributes, and the weighted sum of their activations
is used as a compliment to $\h_t$ in determining the distribution $\p_t$.
Specifically, the distribution is generated by a linear transform followed by a softmax normalization:
\begin{equation}
\label{eqn:attout}
    \p_t \propto \exp\big(\E^T \W^{Y,h} (\h_t + \mbox{diag}(\w^{Y,A}) \sum_i \beta^i_t \sigma(\E\y^i)) \big) ,
\end{equation}
where $\W^{Y,h} \in \mathbb{R}^{d \times n}$ is the projection matrix and
$\w^{Y,A} \in \mathbb{R}^n$ models the relative importance of visual attributes in each dimension of the RNN state space.
The $\E^T$ term is inspired by the transposed weight sharing trick \cite{mao2015learning}
for parameter reduction.

\subsection{Model learning}

The training data for each image consist of input image features $\vv$, $\{A_i\}$
and output caption words sequence $\{Y_t\}$.
Our goal is to learn all the attention model parameters $\mathbf{\Theta}_{A}= \{\U, \V, \W^{*,*}, \w^{*,*}\}$
jointly with all RNN parameters $\mathbf{\Theta}_{R}$ by minimizing a loss function over training set.
The loss of one training example is defined as the total negative log-likelihood of all the words
combined with regularization terms on attention scores $\{\alpha^i_t\}$ and $\{\beta^i_t\}$:
\begin{equation}
\label{eqn:objective}
\min_{\mathbf{\Theta}_A, \mathbf{\Theta}_R} -\sum_t \log p(Y_t) + g(\aalpha) + g(\bbeta),
\end{equation}
where $\aalpha$ and $\bbeta$ are attention score matrices with their $(t, i)$-th entries being $\alpha^i_t$ and $\beta^i_t$.
The regularization function $g$ is used to enforce the completeness of attention paid to every attribute in $\{A_i\}$
as well as the sparsity of attention at any particular time step.
This is done by minimizing the following matrix norms of $\aalpha$ (same for $\bbeta$):
\begin{align}
\label{eqn:regularize}
g(\aalpha) = & \|\aalpha\|_{1,p} + \|\aalpha^T\|_{q,1} \nonumber \\
             = & [\sum_i [ \sum_t \alpha^i_t]^p]^{1/p} + \sum_t [\sum_i (\alpha^i_t)^q]^{1/q} ,
\end{align}
where the first term with $p{>}1$ penalizes excessive attention paid to any single attribute $A_i$
accumulated over the entire sentence, and the second term with $0{<}q{<}1$ penalizes diverted attention to
multiple attributes at any particular time.
We use a stochastic gradient descent algorithm with an adaptive learning
rate to optimize Eq. \eqref{eqn:objective}.

\section{Visual attribute prediction}
\label{sec:attribute}

\begin{figure}[t]
\begin{center}
\includegraphics[width=0.475\textwidth]{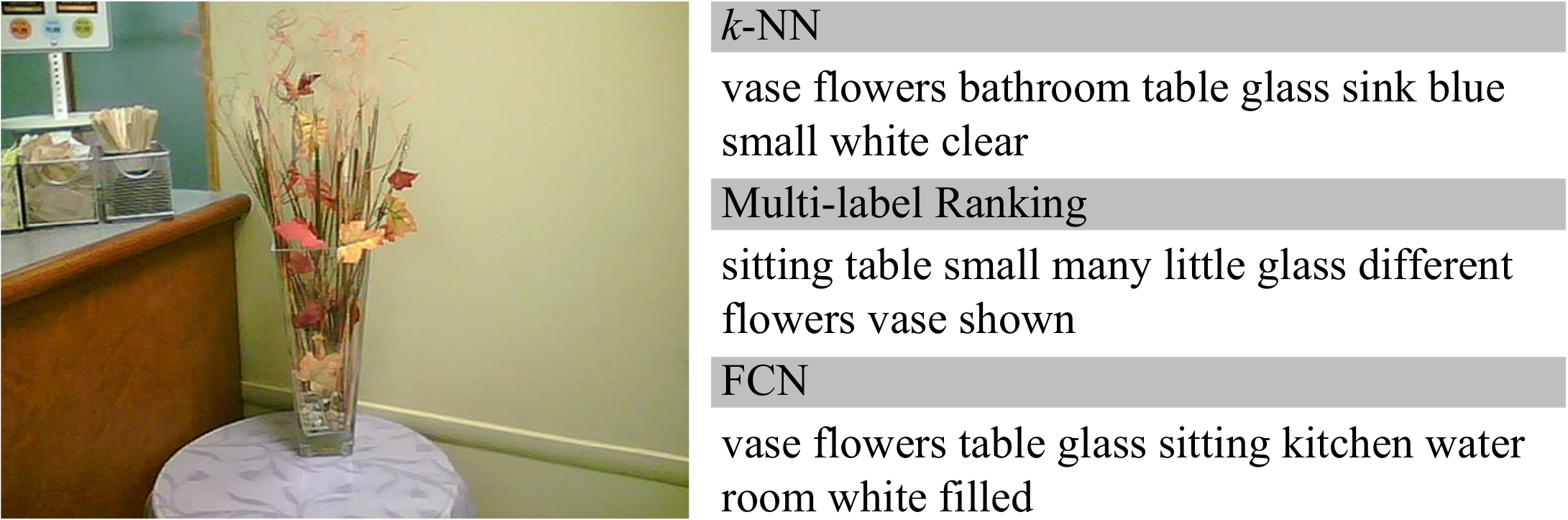}
\end{center}
\caption{An example of top 10 detected visual attributes on an image using different approaches.}
\vspace{-1mm}
\label{fig:example:img}
\vspace{-3mm}
\end{figure}

The prediction of visual attributes $\{A_i\}$ is a key component of our model in both training and testing.
We propose two approaches for predicting attributes from an input image.
First, we explore a non-parametric method based on nearest neighbor image retrieval
from a large collection of images with rich and unstructured textual metadata such as tags and captions.
The attributes for a query image can be obtained by transferring the text information from
the retrieved images with similar visual appearances.
The second approach is to directly predict visual attributes from the input image using a parametric model.
This is motivated by the recent success of deep learning models on visual recognition tasks~\cite{escorcia2015relationship,krizhevsky2012imagenet}.
The unique challenge for attribute detection is that usually there are more than one visual concepts presented in an image,
and therefore we are faced with a multi-label problem instead of a multi-class problem.
Note that the two approaches to obtain attributes are complementary to each other and can be used jointly.
\figurename~\ref{fig:example:img} shows an example of visual attributes predicted for an image using different methods.

\subsection{Non-parametric attribute prediction}

Thanks to the popularity of social media, there is a growing number of images with weak labels, tags, titles and descriptions available on Internet. It has been shown that these weakly annotated images can be exploited to learn visual concepts~\cite{zhou2014conceptlearner}, text-image embedding~\cite{gong2014improving} and image captions~\cite{devlin2015exploring}. One of the fundamental assumptions is that similar images are likely to share similar and correlated annotations. Therefore, it is possible to discover useful annotations and descriptions from visual neighbors in a large-scale image dataset.

We extract key words as the visual attributes for our model from a large image dataset.
For fair comparison with other existing work, we only do nearest neighbor search on our training set to retrieve similar ones to test images.
It is expected that the attribute prediction accuracy can be further improved by using a larger web-scale database.
We use the GoogleNet feature \cite{szegedy2014going} to evaluate image distances, and
employ simple Term-Frequency (TF) to select the most frequent words in the ground-truth captions of the retrieved training images.
In this way, we are able to build a list of words for each image as the detected visual attributes.

\subsection{Parametric attribute prediction}
\label{sec:vcselect}

In addition to retrieved attributes, we also train parametric models to extract visual attributes.
We first build a set of fixed visual attributes by selecting the most common words from the captions in the training data.
The resulting attributes are treated as a set of predefined categories and can be learned
as in a conventional classification problem.

The advance of deep learning has enabled image analysis to go beyond the category level.
In this paper we mainly investigate two state-of-the-art deep learning models for attribute prediction:
using a ranking loss as objective function to learn a multi-label classifier as in \cite{gong2013deep},
and using a Fully Convolutional Network (FCN)~\cite{long2014fully} to learn attributes from local patches as in \cite{fang2014captions}.
Both two methods produce a relevance score between an image and a visual attribute,
which can be used to select the top ranked attributes as input to our captioning model.
Alternatives may exist which can potentially yield better results than the above two models, which is not in the scope of this work.

\section{Experiments}
We perform extensive experiments to evaluate the proposed models. We report all the results using Microsoft COCO caption evaluation tool\footnote{\url{https://github.com/tylin/coco-caption}}, including BLEU, Meteor, Rouge-L and CIDEr~\cite{chen2015microsoft}. We will first briefly discuss the datasets and settings used in the experiments. Next, we compare and analyze the results of the proposed model with other state-of-the-art models on image captioning.

\subsection{Datasets and settings}
\label{sec:exp:impl}
We choose the popular Flickr30k and MS-COCO to evaluate the performance of our models. Flickr30k has a total of $31,783$ images. MS-COCO is more challenging, which has $123,287$ images. Each image is given at least five captions by different
AMT workers. To make the results comparable to others, we use the publicly available splits\footnote{\url{https://github.com/karpathy/neuraltalk}} of training, testing and validating sets for both Flickr30k and MS-COCO. We also follow the publicly available code~\cite{karpathy2014deep} to preprocess the captions (\ie building dictionaries, tokenizing the captions).

\begin{table*}[!t]
\begin{center}
\begin{tabular}{|l|l|*{7}{c|}}
\hline
Dataset & Model & B-1 & B-2 & B-3 & B-4 & METEOR &ROUGE-L & CIDEr \\
\hline \hline
\multicolumn{1}{|c|}{\multirow{5}{*}{ Flickr30k }}
& Ours-GT-ATT & 0.824 & 0.679 & 0.534 & 0.412 & 0.269 & 0.588 & 0.949\\
& Ours-GT-MAX & 0.719 & 0.542 & 0.396 & 0.283 & 0.230 & 0.529 & 0.747\\
& Ours-GT-CON & 0.708 & 0.534 & 0.388 & 0.276 & 0.222 & 0.516 & 0.685\\
\cline{2-9}
& Google NIC~\cite{vinyals2014show} & 0.663 & 0.423 & 0.277 & 0.183& -- & -- & --\\
& Toronto~\cite{xu2015show} & 0.669 & 0.439 & 0.296 & 0.199 & 0.185 & -- & -- \\
\hline
\multicolumn{1}{|c|}{\multirow{5}{*}{ MS-COCO }}
& Ours-GT-ATT & 0.910 & 0.786 & 0.654 & 0.534 & 0.341 & 0.667 & 1.685\\
& Ours-GT-MAX & 0.790 & 0.635 & 0.494 & 0.379 & 0.279 & 0.580 & 1.161\\
& Ours-GT-CON & 0.766 & 0.617 & 0.484 & 0.377 & 0.279 & 0.582 & 1.237\\
\cline{2-9}
& Google NIC~\cite{vinyals2014show} & 0.666 & 0.451 & 0.304 & 0.203& -- & -- & --\\
& Toronto~\cite{xu2015show} & 0.718 & 0.504 & 0.357 & 0.250 & 0.230 & -- & -- \\
\hline
\end{tabular}
\end{center}
\caption{Performance of the proposed models using the ground-truth visual attributes on MS-COCO and Flickr30k. (--) indicates unknown scores.}
\label{tab:coco:gt}
\end{table*}


Our captioning system is implemented based on a Long Short-Term Memory (LSTM) network \cite{vinyals2014show}.
We set $n=m=512$ for the input and hidden layers, and use $\tanh$ as nonlinear activation function $\sigma$.
We use Glove feature representation~\cite{pennington2014glove} with $d=300$ dimensions
as our word embedding $\E$.

The image feature $\vv$ is extracted from the last $1024$-dimensional convolutional layer of
the GoogleNet~\cite{szegedy2014going} CNN model.
Our attribute detectors are trained for the same set of visual concepts as in \cite{fang2014captions} for Microsoft COCO data set. We build and train another independent set of attribute detectors for Flickr30k following the steps in \cite{fang2014captions} using the training split of Flickr30k.
The top $10$ attributes with highest detection scores are selected to form the set $\{A_i\}$ in our best attention model setting.
An attribute set of such size can maintain a good tradeoff between precision and recall.

In training, we use RMSProp~\cite{rmsprop2012} algorithm to do model updating with a mini-batch size of $256$.
The regularization parameters are set as $p=2, q=0.5$ in \eqref{eqn:regularize}.

In testing, a caption is formed by drawing words from RNN until a special end word is reached.
All our results are obtained with the ensemble of 5 identical models trained with different initializations,
which is a common strategy adopted in other work~\cite{vinyals2014show}.

In the following experiments, we evaluate different ways to obtain visual attributes as described in Section~\ref{sec:attribute},
including one non-parametric method ($k$-NN) and two parametric models trained with ranking-loss (RK) and fully-connected network (FCN).
Besides the attention model (ATT) described in Section~\ref{sec:att}, two fusion-based methods to utilize the detected attributes $\{A_i\}$
are tested by simply taking the element-wise max (MAX) or concatenation (CON) of the embedded attribute vectors $\{\E\y^i\}$.
The combined attribute vector is used in the same framework and applied at each time step.

\begin{table*}
\begin{center}
\begin{tabular}{|l|*{10}{c|}}
 \hline
& \multicolumn{5}{|c|}{Flickr30k} &\multicolumn{5}{|c|}{MS-COCO} \\
\hline
Model & B-1 & B-2 & B-3 & B-4 & METEOR  & B-1 & B-2 & B-3 & B-4 & METEOR \\
\hline
Google NIC~\cite{vinyals2014show} & 0.663 & 0.423 & 0.277 & 0.183& --& 0.666 & 0.451 & 0.304 & 0.203& --  \\
m-RNN~\cite{mao2014deep} & 0.60 & 0.41 & 0.28 & 0.19 & -- & 0.67 & 0.49 & 0.35 & 0.25& -- \\
LRCN~\cite{donahue2014long} & 0.587 & 0.39 & 0.25 & 0.165 & -- & 0.628 & 0.442 & 0.304 & 0.21 & --\\
MSR/CMU~\cite{chen2014learning} & -- & -- & -- & 0.126 & 0.164 & -- & -- & -- & 0.19 & 0.204\\
Toronto~\cite{xu2015show} & \textbf{0.669} & 0.439 & 0.296 & 0.199 & 0.185 & \textbf{0.718} & 0.504 & 0.357 & 0.250 & 0.230 \\
\hline
Ours-CON-$k$-NN & 0.619 & 0.426 & 0.291&0.197&0.179 & 0.675 & 0.503 & 0.373 & 0.279 & 0.227\\
Ours-CON-RK & 0.623 & 0.432 & 0.295 & 0.200 & 0.179 & 0.647 & 0.472 & 0.338 & 0.237 & 0.204\\
Ours-CON-FCN & 0.639 & 0.447 & 0.309 & 0.213 & 0.188 & 0.700 & 0.532 &0.398 &0.300 & 0.238\\
\hline
Ours-MAX-$k$-NN & 0.622 & 0.426 & 0.287 & 0.193 & 0.178& 0.673 & 0.501 & 0.371& 0.279&0.227\\
Ours-MAX-RK &0.623 & 0.429 & 0.294 & 0.202 & 0.178 & 0.655 & 0.478 & 0.344 &0.245 &0.208\\
Ours-MAX-FCN & 0.633 & 0.444 &0.306 & 0.21 & 0.181& 0.699 & 0.530 & 0.398 & 0.301 & 0.240\\
\hline
Ours-ATT-$k$-NN & 0.618 & 0.428 & 0.290 & 0.195 & 0.172& 0.676 & 0.505 & 0.375 & 0.281 & 0.227  \\
Ours-ATT-RK & 0.617 & 0.424 & 0.286 & 0.193 & 0.177& 0.679 & 0.506 & 0.375 & 0.282 & 0.231\\
Ours-ATT-FCN & 0.647 & \textbf{0.460} & \textbf{0.324} & \textbf{0.230} & \textbf{0.189}& 0.709 & \textbf{0.537} & \textbf{0.402} & \textbf{0.304} & \textbf{0.243}  \\
\hline
\end{tabular}
\end{center}
\caption{Performance in terms of BLEU-1,2,3,4 and METER compared with other state-of-the-art methods. For those competing methods, we extract their performance from their latest version of paper. The numbers in bold face are the best known results and (--) indicates unknown scores.}
\label{tab:coco:val}
\vspace{-3mm}
\end{table*}

\subsection{Performance on MS-COCO}

Note that the overall captioning performance will be affected by the employed visual attributes generation method. Therefore, we first assume ground truth visual attributes are given and evaluate different ways (CON, MAX, ATT) to select these attributes. This will indicate the performance limit of exploiting visual attributes for captioning. To be more specific, we select the most common words as visual attributes from their ground-truth captions to help the generation of captions.
\tablename~\ref{tab:coco:gt} shows the performance of the three models using the \emph{ground-truth} visual attributes. These results can be considered as the \emph{upper bound} of the proposed models
, which suggest that all of the proposed models (ATT, MAX and CON) can significantly improve the performance of image captioning system, if given visual attributes of high quality.

Now we evaluate the complete pipeline with both attribute detection and selection. The right half of \tablename~\ref{tab:coco:val} shows the performance of the proposed model on the validation set of MS-COCO. In particular, our proposed attention model outperforms all the other state-of-the-art methods in most of the metrics, which are commonly used together for fair and overall performance measurement. Note that B-1 is related to single word accuracy, the performance gap of B-1 between our model and ~\cite{xu2015show} may be due to different preprocessing for word vocabularies.

In Table 2, the entries with prefix ``Ours" show the performance of our method configured with different combinations of attribute detection and selection methods. In general, attention model ATT with attributes predicted by FCN model yields better performance than other combinations over all benchmarks.

For attribute fusion methods MAX and CON, we find using the top $3$ attributes gives the best performance.
Due to the lack of attention scheme, too many keywords may increase the parameters for CON and may reduce the distinction among different groups of keywords for MAX. Both models have comparable performance. The results also suggest that FCN gives more robust visual attributes. MAX and CON can also outperform the state-of-the-art models in most evaluation metrics using visual attributes predicted by FCN. Attention models (ATT) on FCN visual attributes show the best performance among all the proposed models. On the other hand, visual attributes predicted by ranking loss (RK) based model seem to have even worse performance than $k$-NN. This is possible due to the lack of local features in training the ranking loss based attribute detectors.
\setlength{\tabcolsep}{0.45em}
\begin{table*}
\small
\begin{center}
\tabcolsep=0.11cm
\begin{tabular}{ |l|*{15}{c|} }
\hline
\multicolumn{1}{|c|}{\multirow{2}{*}{ Alg }} & \multicolumn{2}{|c|}{B-1} &
\multicolumn{2}{|c|}{B-2} & \multicolumn{2}{|c|}{B-3} &
\multicolumn{2}{|c|}{B-4} & \multicolumn{2}{|c|}{METEOR} & \multicolumn{2}{|c|}{ROUGE-L} & \multicolumn{2}{|c|}{CIDEr} \\
\cline{2-15}
& \multicolumn{1}{|c|}{c5} & \multicolumn{1}{|c|}{c40} & \multicolumn{1}{|c|}{c5} & \multicolumn{1}{|c|}{c40}& \multicolumn{1}{|c|}{c5} & \multicolumn{1}{|c|}{c40} & \multicolumn{1}{|c|}{c5} & \multicolumn{1}{|c|}{c40} & \multicolumn{1}{|c|}{c5} & \multicolumn{1}{|c|}{c40}  & \multicolumn{1}{|c|}{c5} & \multicolumn{1}{|c|}{c40}  & \multicolumn{1}{|c|}{c5} & \multicolumn{1}{|c|}{c40} \\
\hline \hline
ATT & $0.731_{\bm{\textcolor{red}{1}}}$& $0.9_{{2}}$ & $0.565_{\bm{\textcolor{red}{1}}}$ & $0.815_{{2}}$ & $0.424_{\bm{\textcolor{red}{1}}}$  & $0.709_{{2}}$ & $0.316_{\bm{\textcolor{red}{1}}}$ & $0.599_{{2}}$  & $0.250_{{3}}$ & $0.335_{{4}}$ & $0.535_{\bm{\textcolor{red}{1}}}$  & $0.682_{\bm{\textcolor{red}{1}}}$& $0.943_{\bm{\textcolor{red}{1}}}$ & $0.958_{\bm{\textcolor{red}{1}}}$\\
\hline
OV & $0.713_{{6}}$& $0.895_{{3}}$ & $0.542_{{6}}$ & $0.802_{{4}}$ & $0.407_{{4}}$  & $0.694_{{4}}$ & $0.309_{{2}}$ & $0.587_{{3}}$  & $0.254_{\bm{\textcolor{red}{1}}}$ & $0.346_{\bm{\textcolor{red}{1}}}$ & $0.530_{{2}}$  & $0.682_{\bm{\textcolor{red}{1}}}$& $0.943_{\bm{\textcolor{red}{1}}}$ & $0.946_{{2}}$\\
\hline
MSR Cap & $0.715_{{5}}$& $0.907_{\bm{\textcolor{red}{1}}}$ & $0.543_{5}$ & $0.819_{\bm{\textcolor{red}{1}}}$ & $0.407_{{4}}$  & $0.710_{\bm{\textcolor{red}{1}}}$ & $0.308_{{3}}$ & $0.601_{\bm{\textcolor{red}{1}}}$  & $0.248_{{4}}$ & $0.339_{{2}}$ & $0.526_{{4}}$  & $0.680_{{3}}$& $0.931_{{3}}$ & $0.937_{{3}}$\\
\hline
mRNN & $0.716_{{4}}$ & $0.890_6$ & $0.545_4$ & $ 0.798_6$ & $0.404_6$ & $0.687_6$ & $0.299_6$ & $0.575_6$ & $0.242_9$ & $0.325_8$ & $0.521_6$ & $0.666_6$ & $0.917_4$
& $0.935_4$ \\
\hline
\end{tabular}
\end{center}
\caption{Performance of the proposed attention model on the online MS-COCO testing server (\url{https://www.codalab.org/competitions/3221\#results}), comparing with other three leading methods. The subscripts indicate the current ranking of the individual algorithms with respect to the evaluation metrics. ATT refers to our entry, OV refers to the entry of OriolVinyals, MSR Cap refers to MSR Captivator, and mRNN refers to mRNN\_share.JMao.}
\vspace{-3mm}
\label{tab:att:testing:server}
\end{table*}

\textbf{Performance on MS-COCO 2014 test server}
We also evaluate our best model, Ours-ATT-FCN, on the MS COCO Image Captioning Challenge sets c5 and c40 by uploading results to the official test server. In this way, we could compare our method to all the latest state-of-the-art methods. Despite the popularity of this contest, our method has held the top 1 position by many metrics at the time of submission. \tablename~\ref{tab:att:testing:server} lists the performance of our model and other leading methods. Besides the absolute scores, we provide the rank of our model among all competing methods for each metric. By comparing with two other leading methods, we can see that our method achieves better ranking across different metrics. All the results are up-to-date at time of submission.
\begin{figure*}[!htbp]
\begin{center}
\subfigure[]{
\includegraphics[width=0.235\textwidth]{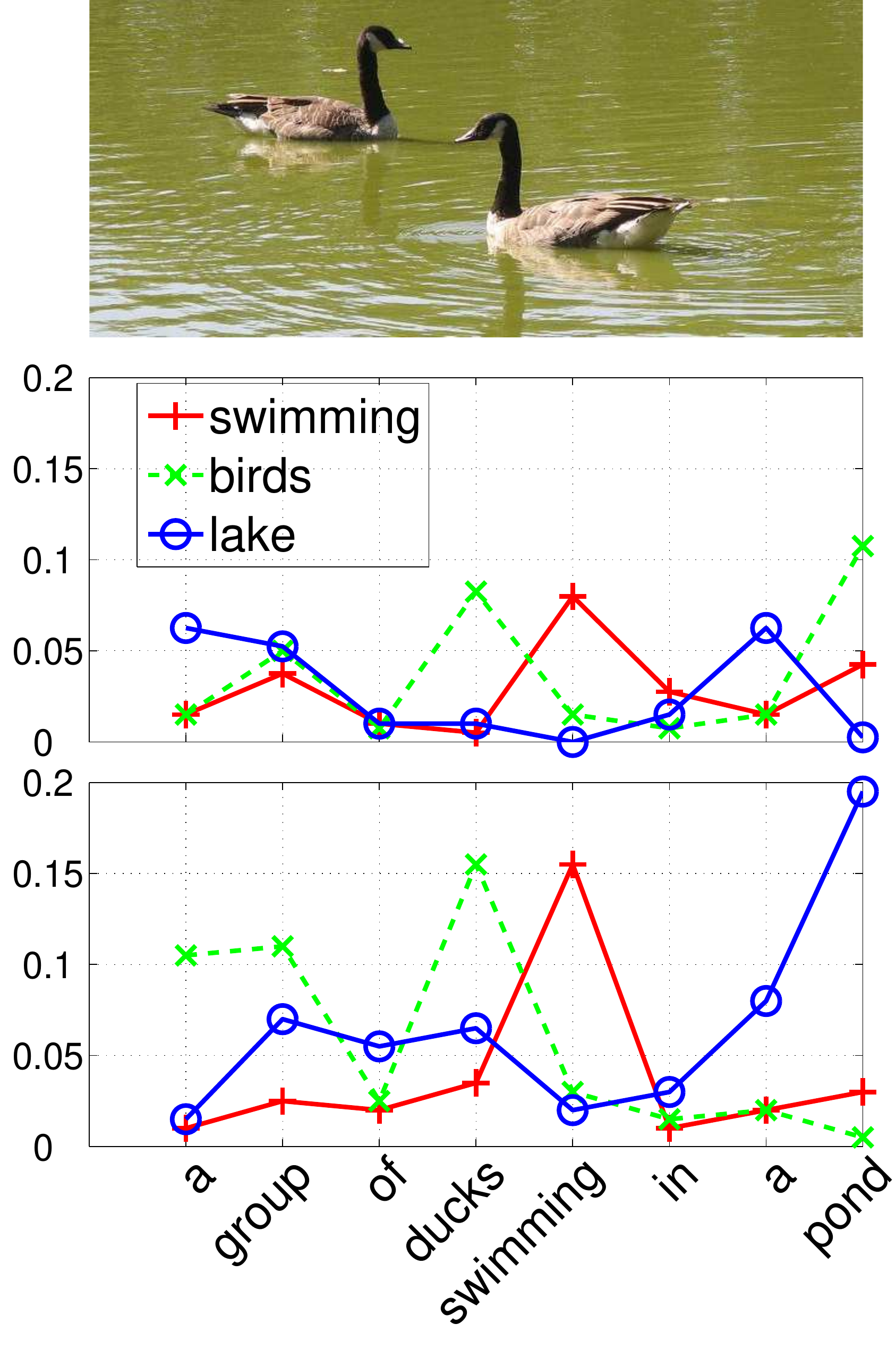}
\label{fig:att:1}
}
\subfigure[]{
\includegraphics[width=0.235\textwidth]{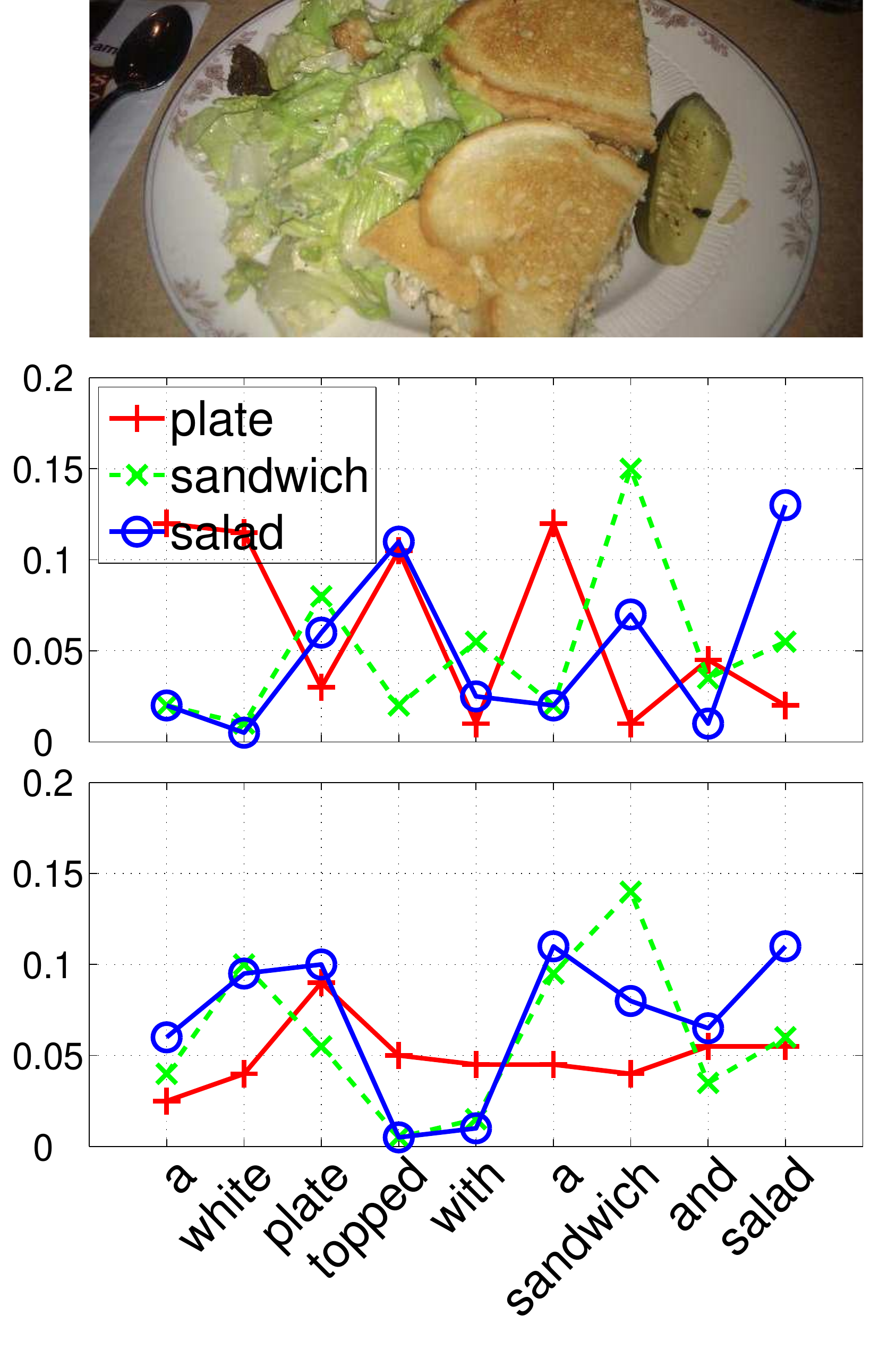}
\label{fig:att:2}
}
\subfigure[]{
\includegraphics[width=0.235\textwidth]{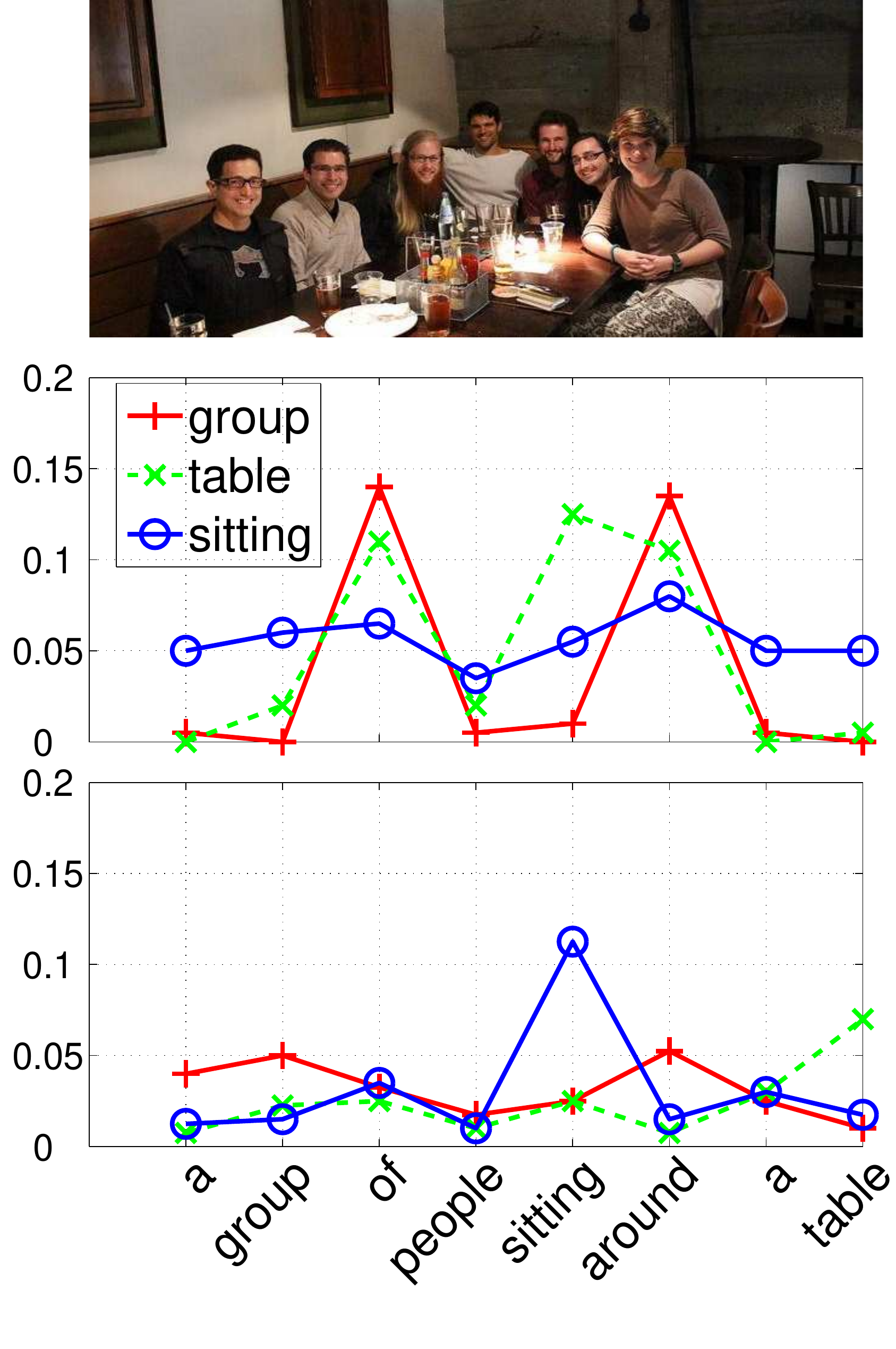}
\label{fig:att:3}
}
\subfigure[]{
\includegraphics[width=0.235\textwidth]{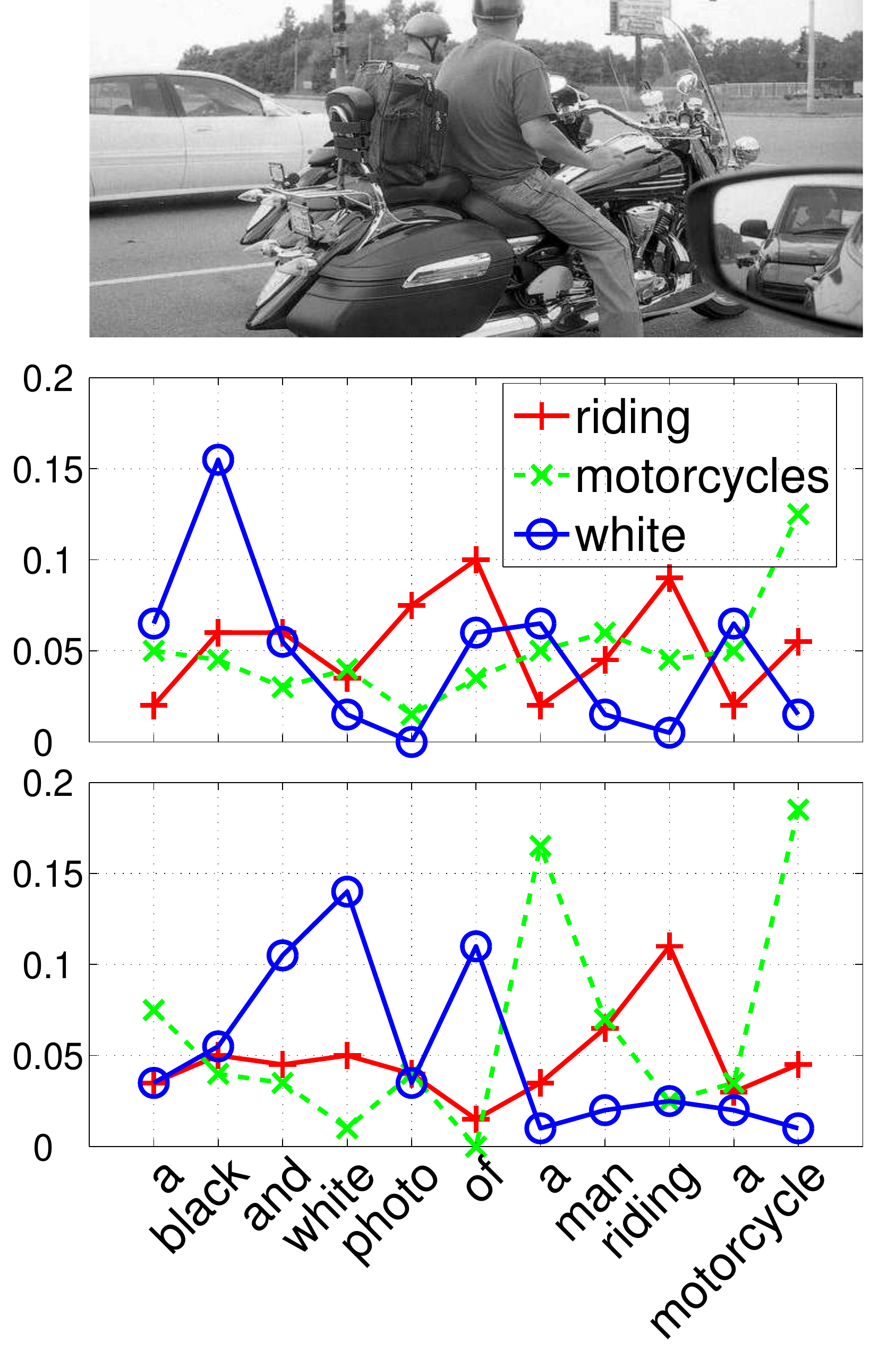}
\label{fig:att:4}
}
\end{center}
\vspace{-5mm}
\caption{Examples of attention weights changes along with the generation of captions. \textbf{Second row}: input attention weights $\alpha$. \textbf{Third row}: output attention weights $\beta$. The $X$-axis shows the generated caption for each image and the $Y$-axis is the weight. We only show the change of weights on three top visual attributes for each example.}
\label{fig:att}
\end{figure*}

\subsection{Performance on Flickr30k}
We now report the performance on Flickr30k dataset. Similarly, we first train and test our models by using the ground-truth visual attributes to get an upper-bound  performance. The obtained results are listed in \tablename~\ref{tab:coco:gt}. Clearly, with correct visual attributes, our model is able to improve caption results by a large margin comparing to other methods. We then conduct the full evaluation. As shown in the left half of \tablename~\ref{tab:coco:val}, the performance of our models are consistent with that on MS-COCO, and Ours-ATT-FCN achieves significantly better results over all competing methods in all metrics, except B-1 score, for which we have discussed potential causes in previous section.

\begin{figure*}[!htbp]
\begin{center}
\includegraphics[width=0.95\textwidth]{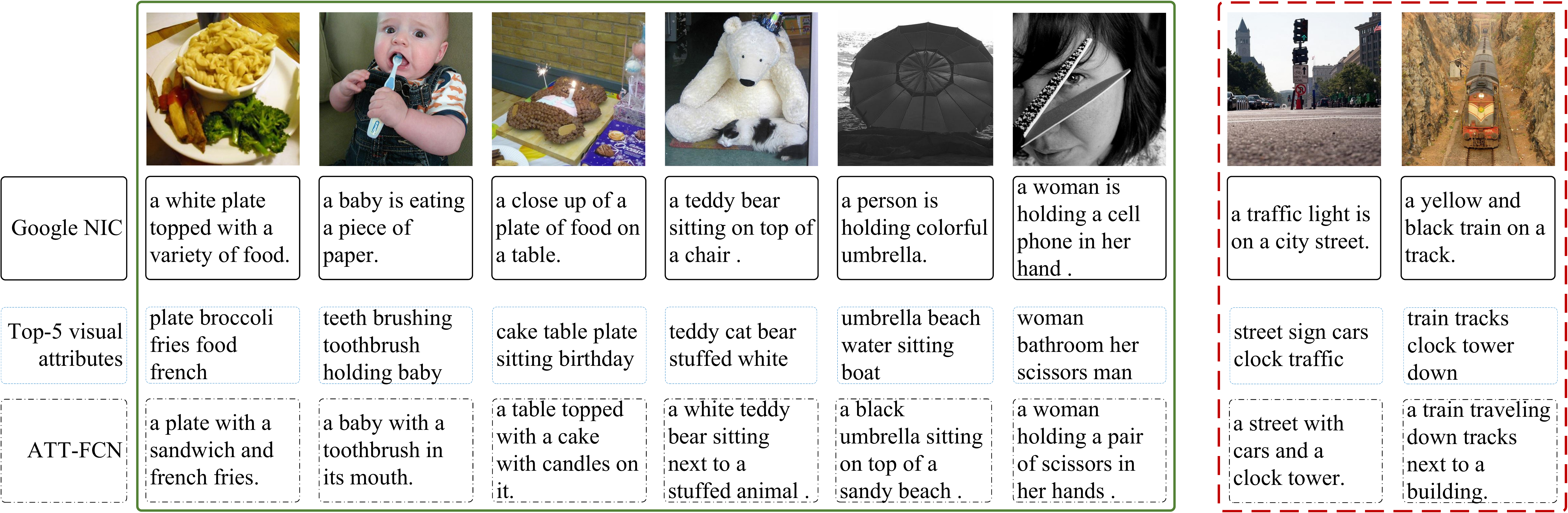}
\end{center}
\vspace{-2mm}
\caption{Qualitative analysis on impact of visual attributes. The left six examples (green solid box) shows that the visual attributes help generate more accurate captions. The right two examples (red dashed box) indicate that incorrect visual attributes may mislead the model.}
\label{fig:va:analysis}
\vspace{-4mm}
\end{figure*}
\subsection{Visualization of attended attributes}
We now provide some representative captioning examples in \figurename~\ref{fig:att} for better understanding of our model. For each example, \figurename~\ref{fig:att} contains the generated captions for several images with the input attention weights $\alpha_t^i$ and the output attention weights $\beta_t^i$ plotted at each time step. The generated caption sentences are shown under the horizontal time axis of the curve plots, and each word is positioned at the time step it is generated. For visual simplicity, we only show the attention weights of top attributes from the generated sentence. As captions are being generated, the attention weights at both input and output layers vary properly as sentence context changes, while the distinction between their weights shows the underlying attention mechanisms are different. In general, the activations of both $\alpha$ and $\beta$ have strong correlation with the words generated. For example, in the \figurename~\ref{fig:att:1}, the attention on ``swimming'' peaks after ``ducks'' is generated for both $\alpha$ and $\beta$.
In \figurename~\ref{fig:att:4}, the concept of ``motorcycle'' attracts strong attention for both $\alpha$ and $\beta$. The $\beta$ peaks twice during the captioning process, one after ``photo of'' and the other after ``riding a'', and both peaks reasonably align with current contexts.
It is also observed that, as the output attention weight, $\beta$ correlates with output words more closely; while the input weights $\alpha$ are allocated more on background context such as the ``plate'' in \figurename~\ref{fig:att:2} and the ``group'' in \figurename~\ref{fig:att:3}. This temporal analysis offers an intuitive perspective on our visual attributes attention model.
\begin{table}
\begin{center}
\begin{tabular}{|l|*{7}{c|}}
\hline
Alg & B-1 & B-2 & B-3 & B-4 & MT &RG & CD \\
\hline
\hline
Input & 0.88 & 0.75 & 0.62 & 0.50 & 0.33 & 0.65 & 1.56 \\
Output & 0.89 & 0.76 & 0.62 & 0.50 & 0.33 & 0.65 & 1.58 \\
Full & 0.91 & 0.79 & 0.65 & 0.53 & 0.34 & 0.67 & 1.68 \\
\hline
\end{tabular}
\end{center}
\caption{The performance of different models with input attention (first row), output attention (second row), and both attentions (third row) using the ground-truth visual attributes on MS-COCO validation dataset. We use abbreviations MT, RG and CD to stand for METEOR, ROUGE-L and CIDEr respectively.}
\label{tab:gt:module}
\end{table}
\subsection{Analysis of attention model}
As described in Section~\ref{sec:attin} and Section~\ref{sec:attout}, our framework employs attention at both input and output layers to the RNN module. We evaluate the effect of each of the individual attention modules on the final performance by turning off one of the attention modules while keeping the other one in our ATT-FCN model. The two model variants are trained on MS-COCO dataset using the ground-truth visual attributes, and compared in \tablename~\ref{tab:gt:module}. The performance of using output attention is slightly better than only using input attention on some metrics. However, the combination of this two attentions improves the performance by several percents on almost every metric.
This can be attributed to that fact that attention mechanisms at input and output layers are not the same, and each of them attend to different aspects of visual attributes. Therefore, combining them may help provide a richer interpretation of the context and thus lead to improved performance.
\subsection{The role of visual attributes}
We also conduct a qualitative analysis on the role of visual attributes in caption generation. We compare our attention model (ATT) with Google NIC, which corresponds to the LSTM model used in our framework. \figurename~\ref{fig:va:analysis} shows several examples. We can find that visual attributes can help our model to generate better captions, as shown by the examples in the green box. However, irrelevant visual attributes may disrupt the model to attend on incorrect concepts. For example, in the left example in the red dashed box, ``clock'' distracts our model to the clock tower in background from the main objects in foreground. In the rightmost example, and ``tower'' may be the culprit of the word ``building'' in the predicted caption.
\section{Conclusion}
\label{sec:conclusion}

In this work, we proposed a novel method for the task of image captioning, which achieves state-of-the-art performance across popular standard benchmarks. Different from previous work, our method combines top-down and bottom-up strategies to extract richer information from an image, and couples them with a RNN that can selectively attend on rich semantic attributes detected from the image. Our method, therefore, exploits not only an overview understanding of input image, but also abundant fine-grain visual semantic aspects. The real power of our model lies in its ability to attend on these aspects and seamlessly fuse global and local information for better caption. For next steps, we plan to experiment with phrase-based visual attribute with its distributed representations, as well as exploring new models for our proposed semantic attention mechanism.


\section*{Acknowledgment}
This work was generously supported in part by Adobe Research and New York State through the Goergen Institute for Data Science at the University of Rochester.
{\small
\bibliographystyle{ieee}
\bibliography{egbib}
}

\end{document}